\documentclass{article}
\usepackage{spconf,amsmath,epsfig}
\begin{document}\sloppy

\def\x{{\mathbf x}}
\def\L{{\cal L}}

\title{A Comprehensive Study of Sparse Codes on Abnormality Detection}

\name{Huamin Ren$^1$, Hong Pan$^2$, S\o ren Ingvor Olsen$^3$, Thomas B. Moeslund$^4$}
\address{$^{1,4}$Department of Architecture, Design and Media Technology, Aalborg University, Denmark\\
$^{2,3}$Department of Computer Science, University of Copenhagen, Denmark\\
hr@create.aau.dk}

\maketitle

\begin{abstract}
Sparse representation has been applied successfully in abnormal event detection, in which the baseline is to learn a dictionary accompanied by sparse codes. While much emphasis is put on discriminative dictionary construction, there are no comparative studies of sparse codes regarding abnormality detection. We comprehensively study two types of sparse codes solutions - greedy algorithms and convex L1-norm solutions - and their impact on abnormality detection performance. We also propose our framework of combining sparse codes with different detection methods. Our comparative experiments are carried out from various angles to better understand the applicability of sparse codes, including computation time, reconstruction error, sparsity, detection accuracy, and their performance combining various detection methods. Experiments show that combining OMP codes with maximum coordinate detection could achieve state-of-the-art performance on the UCSD dataset~\cite{Mahadevan2010}.
\end{abstract}
%

\section {Introduction}
\label{Introduction}

Sparse representation has gained a great deal of attention since being applied effectively in many image analysis applications, such as image denoising~\cite{donoho2002optimally} and action recognition~\cite{JiangTPAMI}. Sparse representation finds the most compact representation of a signal in terms of linear combination of atoms in an overcomplete dictionary. As is pointed out in~\cite{Huang06sparserepresentation}, research has focused on three aspects of sparse representation: pursuit methods for solving the optimization problem, such as matching pursuit~\cite{MP1993}, orthogonal matching pursuit~\cite{OMP1993}, and basis pursuit~\cite{Chen98atomicdecomposition}; dictionary design, such as the K-SVD~\cite{Aharon05k} method and the BSD algorithm~\cite{RenBMVC2015}; and the applications of the sparse representation for different tasks, such as abnormal event detection. Abnormal event detection is the core of video surveillance applications, which could assist people in various situations, such as monitoring patients/children, observing people and vehicles within a busy environment, or preventing theft and robbery. The aim of the method is to learn normal patterns or behaviors through training and detect any abnormal or suspicious behaviors in test videos. 

Research on sparse representation can be generally divided into dictionary learning~\cite{Lu13} and sparse coding~\cite{MP1993}~\cite{OMP1993}~\cite{Chen98atomicdecomposition}. Dictionary learning aims to obtain atoms (or basis vectors) for a dictionary. Such atoms could be either predefined, e.g., undecimated Wavelets, steerable Wavelets, Contourlets, Curvelets, and more variants of Wavelets, or learned from the data itself. Sparse coding, on the other hand, attempts to find sparse codes (or coefficients) by giving a dictionary, i.e., finding the solution to the underdetermined system of equations $y = Dx$ either by  greedy algorithms or convex algorithms. Through sparse coding, input features can be approximately represented as a weighted linear combination of a small number of (unknown) basis vectors. 

When applying sparse representation on abnormal event detection, much emphasis is put on dictionary learning. A common procedure is: first, visual features are extracted either on a spatial or temporal domain. A dictionary $D$ is then learned based on these visual features, which consists of basis vectors capturing high-level patterns in the input features, as in~\cite{Lu13, RenBMVC2015}. A sparse representation of a feature is a linear combination of a few elements or atoms from a dictionary. Mathematically, it can be expressed as $y = Dx$, where $y \in R^{p}$ is a feature of interest, $D \in R^{p \times m}$ is a dictionary, and $x \in R^{m}$ is the sparse representation of $y$ in $D$. Typically $m \gg p$ results in an overcomplete or redundant dictionary. During the detection procedure, each testing feature can be determined as normal or an anomaly based on its reconstruction error. 

However, most approaches use only an approximate reconstruction error to save computation; for example, the least square error. This means the sparse codes are actually not taken into consideration during the detection. In fact, the impact of sparse codes generated by different approaches is still unclear. Therefore, we offer a comprehensive study of the sparse codes, in terms of their performance on abnormal event detection. Among the huge research of codes representations, we put special attention on two major types:  greedy algorithms and L1-norm minimization algorithms. 

Greedy algorithms rely on an interactive approximation of the feature coefficients and supports, either by iteratively identifying the support of the feature until a convergence criterion is met, or by obtaining an improved estimate of the sparse signal at each iteration that attempts to account for the mismatch with the measured data. Compared to L1-norm minimization methods, greedy algorithms are much faster, and thus are more applicable to very large problems.

Meanwhile, L1-norm minimization has become a popular tool to solve sparse coding, which benefits both from efficient algorithms and a well-developed theory for generalization properties and variable selection consistency~\cite{zhang2009}. We list two common L1-norm minimization formulations in E.q. \ref{eq:l1reg} and E.q. \ref{eq:l1err}. Since the problem is convex, there are efficient and accurate numerical solvers.

\newcommand{\argmin}{\arg\!\min}
\begin{equation}\label{eq:l1reg}
\hat{x} = \argmin_x  \frac{1}{2} \| Dx - y \| _2^2 + \lambda \| x \|_1
\end{equation}

\begin{equation}\label{eq:l1err}
\hat{x} = \argmin_x \| x \|_1 \quad \text{subject to} \quad \| Dx - y \| _2  \leq \epsilon
\end{equation}

Our main contributions are: 1) we offer a comprehensive study of sparse codes, in terms of their reconstruction error, sparsity, computation time and detection performance on anomaly datasets;  2) we propose a framework to detect abnormality, which combines sparse representation with various detection methods; and 3) we provide insights into the impact of sparse representation and their detection methods. 


The remainder of this paper is organized as follows. We give a brief review of greedy algorithms and L1-norm solutions in Sec.2 and propose our framework of abnormal event detection in Sec.3, which combines sparse codes with various detection methods. We show our comparative results in Sec.4 and concludes the paper with discussions and future work in Sec.5.

\section{Sparse Codes Representation}\label{sec:pre}
There are various ways of generating sparse codes through optimization solutions. We introduce two categorized solutions: greedy algorithms and L1-norm approximation solutions.

\subsection{Greedy Algorithms}\label{sec:greedy}
We review two broad categories of greedy methods to reconstruct $y$, which are called `greedy pursuits' and `threshold' algorithms. Greedy pursuits can be defined as a set of methods that iteratively build up an estimate $x$. They contains three basic steps. First, the $x$ is set to a zero vector. Second, these methods estimate a set of non-zero components of $x$ by iteratively adding new components that are deemed to be non-zeros. Third, the values for all non-zeros components are optimized. In contrast, thresholding algorithms alternate between element selection and element pruning steps.

There is a large and growing family of greedy pursuit methods. The general framework in greedy pursuit techniques is 1) to select an element and 2) to update the coefficients. Matching Pursuit (MP)~\cite{MP1993} discusses a general method for approximate decomposition in E.q. \ref{eq:mp}, which addresses the sparsity issue directly. The algorithm selects one column from $D$ at a time and only the coefficient associated with the selected column is updated at each iteration. More concretely, it starts from an initial approximation $x$ (0) = 0 and residual $R$(0) = $x$, then builds up to a sequence of sparse approximations stepwise. At stage $k$, it identifies the dictionary atom that best correlates with the residual and then adds to the current approximation a scalar multiple of that atom. After $m$ steps, one has a sparse code in E.q. \ref{eq:mp} with residual $R = R^{(m)}$.
\begin{equation}\label{eq:mp}
y = \sum_{i=1}^{m} x_{r_i}d_{r_i} + R^{(m)}
\end{equation}

Orthogonal Matching Pursuit (OMP)~\cite{OMP1993}, updates $x$ in each iteration by projecting $y$ orthogonally onto the columns of $D$ associated with the current support atoms. Different from MP, OMP never reselects an atom and the residual at any iteration is always orthogonal to all currently selected atoms in the dictionary. Another difference is that OMP minimizes the coefficients for all selected atoms at iteration $k$, while MP only updates the coefficient of the most recently selected atom. In order to speed up pursuit algorithms, it is necessary to select multiple atoms at a time; therefore, the algorithms are proposed to keep computational costs low enough for applying to large-scale problems, such as Stagewise Orthogonal Matching Pursuit (StOMP)~\cite{StOMP2012}. These algorithms choose the element that meets some threshold criterion at the atom selection step and has demonstrated both theoretical and empirical effectiveness for the large system. 

Greedy algorithms are easy to implement and use and can be extremely fast. However, they do not have recovery guarantees, i.e., how well each sample can be reconstructed by the dictionary and their sparse codes, compared to L1-norm approximations. 
%

\subsection{L1-norm Approximation}\label{sec:l1}
L1-norm approximation replaces the L0 constraint with a relaxed L1-norm. For example, in the Basis Pursuit method (BP)~\cite{Chen98atomicdecomposition}, an almost everywhere differentiable and often convex cost function is applied, while in the Focal Underdetermined System Solver (FOCUSS) algorithm~\cite{Focuss2001}, a more general model is optimized. 

Donoho and etc.~\cite{donoho2002optimally} sugguest that for some measurement matrices $D$, the generally NP-Hard problem (L0 norm) should be equivalent to its convex relaxation: L1 norm, see E.q. \ref{eq:l1reg}. The convex L1 problem can be solved using methods of linear programming. Representative work includes Basis Pursuit (BP). Instead of seeking sparse representations directly, it seeks representations that minimize the L1 norm of the coefficients. Furthermore, BP can compute sparse solutions in situations where greedy algorithms fail. The Lasso algorithm~\cite{tibshirani96regression} is quite similar to BP and is, in fact, know as Basis Pursuit De-Noising (BPDN) in some areas. Rather than trying to minimize the L1-norm like BP, the Lasso places a restriction on its value.

The FOCUSS algorithm has two integral parts: a low-resolution initial estimate of the real signal and the iteration process that refines the initial estimate to the final localized energy solution. The iterations are based on the weighted norm minimization of the dependent variable with the weights acting as a function of the preceding iterative solutions. The algorithm is presented as a general estimation tool usable across different applications. In general, L1-norm methods offer better performance in many cases, but they are also more demanding with respect to computation.


\section{Sparse Code Based Detection}\label{sec:sp}

\begin{figure}[!htbh]
\vskip 0.2in
\begin{center}
\centerline{\includegraphics[width=\columnwidth]{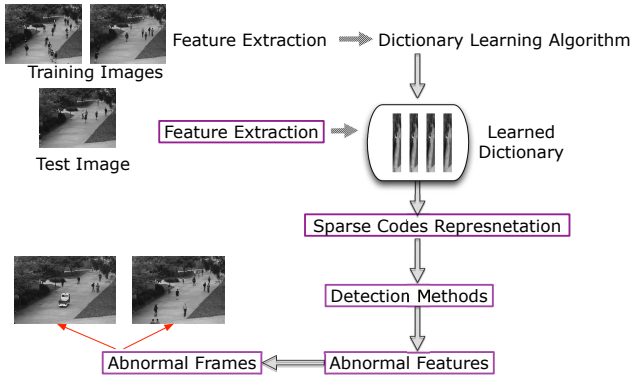}}
\caption{Our framework of combining sparse codes with different detection methods.}
\label{fig:framework}
\end{center}
\vskip -0.2in
\end{figure} 

\noindent In addressing the detection of abnormal behaviors based on sparse codes, two issues should be addressed: 1) how to generate the sparse codes, i.e., the solution of $x$; and 2) how to determine whether the testing code is normal or anomalous. For the first issue, various sparse codes discussed in Sec.2 could be adopted; while for the second issue, we take various detection methods into consideration; our proposed abnormal event detection framework is shown in Fig. \ref{fig:framework}. 

After a testing feature is represented by a sparse code, the detection method determines whether it is normal or abnormal. There are two commonly used detection methods: the reconstruction error (RE) and the approximated reconstruction error (ARE). In terms of sparse codes, the high response of dictionary atoms, or concentrated non-zeros in coefficients, may indicate a connection to a possible normality. Unfortunately, these codes property and their connection with normality or abnormality are not explored yet. Therefore, we also introduce maximum coordinate (MC) and the non-zero concentration (NC) as two new detection methods.

\textbf{Reconstruction Error (RE):} Most existing approaches treat dictionary learning and detection as two separate processes, i.e., a dictionary is typically learned based on the training data, and then different measurements are adopted to determine whether the testing sample is an anomaly. More sophisticated approaches unify these two processes into a mixed reconstructive and discriminative formulation. Nevertheless, a basic measurement that is widely used in both cases is reconstruction error. The reconstruction error of the testing sample $y$, according to the dictionary $D$, is represented as: $\| y - D\alpha \|_2^2$, where $\alpha$ is the sparse code of $y$. 

\textbf{Approximate Reconstruction Error (ARE):} To speed up detection, reconstruction error is sometimes not calculated based on sparse codes through an optimization solution; instead it is approximated by the least squares~\cite{Lu13}; thus, the reconstruction error is calculated as:  $\| y - D(D^TD)^{-1}D^Ty \|$.

\textbf{Maximum Coordinate (MC):} Given a testing sample $y$, its sparse code is denoted as $\alpha$. Ideally, all non-zero entries in the estimate $\alpha$ would be associated with the columns of the dictionary from a normal pattern (note that only normal data is used during the training). Then we could detect $y$ as a normal feature if a single largest entry in $\alpha$ were found; otherwise, it would be detected as an anomaly.

\textbf{Non-zero Concentration (NC):} Inspired by~\cite{RenBMVC2015}, the distribution of non-zeros is more important to the detection than the location of non-zero elements. Thus, we propose a detection measurement called non-zero concentration. Based on the dictionary proposed in~\cite{RenBMVC2015}, a normal code should have a non-zero concentration property, i.e., non-zeros concentrated in the dictionary that has the smallest reconstruction error. Anomalies can be detected if no concentration is found on any of the existing dictionaries.


\section{Experimental Results}\label{sec:experiments}
We provide a comprehensive study the abnormality detection performance of sparse codes. Our experiments are carried out on the UCSD~\cite{Mahadevan2010} Ped1 dataset, due to that it is a popular abnormal event detection dataset, and many detection results are reported. We start by evaluating the performance of various sparse codes, especially comparing sparse codes generated by two types of algorithms: greedy algorithms and L1-norm approximation algorithms. The following aspects are highlighted: computation time, reconstruction error, the ratio of sparsity in codes, and their performance on abnormal event detection. Next, we use the OMP algorithm to generate sparse codes, and then we combine the codes with different detection methods concluding by evaluating their detection performance with state-of-the-art algorithms.

\subsection{Dataset and Settings}
UCSD Ped1 dataset~\cite{Mahadevan2010} is a frequently used public dataset for detecting abnormal behaviors. It includes clips of groups of people walking towards and away from the camera with some perspective distortion. There are 34 training videos and 36 testing videos with a resolution of 238 $\times$ 158. Training videos contain only normal behaviors. Testing videos are abnormal behaviors where there are either non-pedestrian entities in the walkways or anomalous pedestrian motion patterns. 

We use the spatial-temporal cubes, in which 3D gradient features are computed, which mimics the setting in~\cite{KratzN09}. Each frame is divided into patches with a size of 23 $\times$ 15. Consecutive 5 frames are used to form 3D patches, and gradients features are extracted in each patch. See details in~\cite{KratzN09}. Through this, we obtain 500-dimensional visual features and reduce them to 100 dimension by using the PCA algorithm.


\subsection{Comparison of Sparse Codes }

We evaluate sparse codes from four perspectives: computation time, reconstruction error, the ratio of sparsity in codes, and the codes' performance on abnormal event detection based on their reconstruction error.

We randomly select 1\% of the training features (238,000 features in total), use the K-SVD algorithm~\cite{Aharon05k} to construct a dictionary consisting of 1000 atoms, and generate sparse codes by applying various algorithms. There are many algorithms available; we select only representative greedy algorithms (OMP, MP, StOMP) and compare them with representative L1-norm solutions (BP and Lasso algorithm). The reconstruction error is calculated by $Re = \| y - Dx \|_2^2$. We also calculate the mean ratio of sparsity in the codes, i.e., the average percentage of non zeros in the dimension of the codes (1000). We report these results as well as computation time in Tab.~\ref{tab:codes}.  Greedy algorithms need far less time to compute, and the OMP achieves the fastest computation, followed by the StOMP algorithm. OMP is  approx. 180 times faster than the Lasso algorithm. Both OMP and StOMP could achieve sparser solutions, while BP could obtain an extremely dense solution with an exact recovery. 

\begin{table*}[!tb]
\caption{Comparison of greedy algorithms and L1-norm solutions on sparse code generation.}
\label{tab:codes}
\vskip 0.15in
\begin{center}
\begin{small}
\begin{sc}
\begin{tabular}{lcccr}
\hline
Algorithms & Computation Time (s) & Reconstruction Error & Sparsity (\%) \\
\hline
MP    & 166.00 & 0 & 31.8\%\\
OMP & 1.83 &  0.4236 & 1.9\%\\
StOMP    & 15.79& 0 & 10\%\\
BP     &114.20 & 0 & 100\%\\
Lasso     & 333.49 & 0.0005 & 9.9\%\\
\hline
\end{tabular}
\end{sc}
\end{small}
\end{center}
\vskip -0.1in
\end{table*}

\begin{table*}[!tb]
\caption{Comparative results on UCSD Ped1: frame-level evaluation results (AUC and EER) and pixel-level evaluation results (AUC and EDR) are reported.}
\label{tab:auc}
\vskip 0.15in
\begin{center}
\begin{small}
\begin{sc}
\begin{tabular}{lccccr}
\hline
Algorithms & AUC (frame-level) & EER  & AUC (pixel-level) & EDR & Computation Time (s) \\
\hline
MP    & 0.6956	& 0.3547	& 0.3898 & 0.5716 & 13342 \\
OMP & 0.5003 &	0.5052	& 0.2849	& 0.6637	& 527 \\
StOMP    & 0.5415	& 0.465	& 0.3494	& 0.6190	& 4668\\
BP     & 0.5454	& 0.4764 &	0.3057	& 0.6479	& 38949\\
Lasso     &  0.5305 &  0.5173& 0.3132&  0.6383 & 56400 \\
\hline
\end{tabular}
\end{sc}
\end{small}
\end{center}
\vskip -0.1in
\end{table*}

To measure the accuracy of abnormality detection, we calculate the reconstruction error of each feature and register features with large reconstruction errors as anomalies. A frame with an abnormal feature is considered a positive frame. To compare performance, we adopt two popular evaluation criteria in abnormality detection: frame-level evaluation and pixel-level evaluation, which are defined in~~\cite{Mahadevan2010}. We follow precisely their setting in our evaluation, which is to say that in the frame-level evaluation, a frame is considered abnormal if it contains at least one anomaly feature. In contrast, for the pixel-level evaluation, a frame is marked as a correctly detected abnormality if at least 40\% of  the truly abnormal pixels are detected. Ground truth on frame-level and pixel-level annotation is available, and we calculate the true positive and false positive rates to draw ROC curves, and report the Area Under the Curve (AUC). Following~\cite{Mahadevan2010}, we obtain the value when the false positive number equals the missing value. These are called the equal error rate (EER) and equal detected rate (EDR) in the frame and pixel-level evaluations, respectively. See Tab.~\ref{tab:auc} for details. In the frame-level evaluation, the MP algorithm achieves the best results with a moderate computation time. The StOMP algorithm is relatively fast, and the AUC is satisfactory.

It is worth noting that the pixel-level AUC is lower than the frame-level AUC in general because the pixel-level evaluation is stricter and takes location into consideration. In the frame-level evaluation, there could be a coincidental detection - a normal feature could be erroneously detected as an anomaly in an abnormal frame, and this erroneous detection could end up with a correct detection of that frame. In pixel-level evaluation in contrast, a frame is marked as a correctly detected abnormality only if a sufficient number of anomaly features has been found. Compared to the MP algorithm, the StOMP algorithm can achieve a competitive detection result in the pixel-level evaluation, but it is three times faster than the MP algorithm. The BP algorithm also performs well on pixel-level detection; however, its high computation cost hampers its application in real detection problems.

In summary, greedy algorithms compute quickly, but their reconstruction errors are larger than L1-norm solutions. Convex relaxations, such as the BP and the Lasso algorithm, have better theoretical guarantees and recovery ability, but they are more time consuming. Surprisingly, greedy algorithms, especially the StOMP algorithm, seem to perform better on pixel-level detection, which means that they could more accurately localize anomaly features.

\subsection{Comparison of Combining Sparse Codes with Detection Methods }
We choose the OMP algorithm to generate sparse codes due to computation considerations, combine them with four types of detection methods (RE, ARE, MC, NC), and compare their detection performance. We draw comparative frame-level AUC curves that corresponds to the detection methods. Furthermore, we compare these combinations with state-of-the-art methods on abnormality detection.

\begin{figure}[!htbh]
\vskip 0.2in
\begin{center}
\centerline{\includegraphics[width=.95\columnwidth]{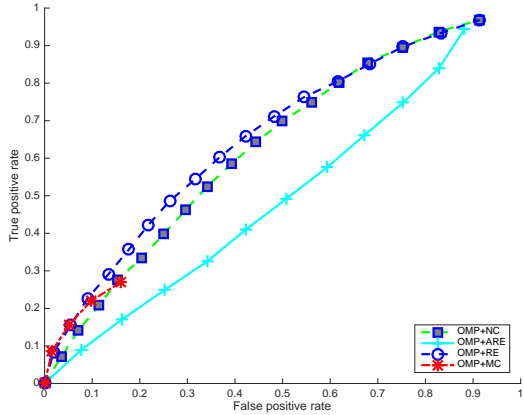}}
\caption{Combining sparse codes generated by the OMP algorithm with various detection methods: non-zero concentration (NC), approximate reconstruction error (ARE), reconstruction error (RE) and maximum coordinate (MC). }
\label{fig:methods}
\end{center}
\vskip -0.2in
\end{figure} 

As displayed in Fig. \ref{fig:methods}, abnormality detection by computing the real reconstruction error outperforms the estimated reconstruction error on frame-level evaluation, which further validates the idea that the decomposition of real coefficients is necessary. Among all of these approaches, OMP+RE achieves the best AUC score on frame-level evaluation (0.6603), followed by MC (0.6340), NC (0.5697) and ARE (0.5013). We give further insight into how accurate the detection is in an even stricter pixel-level evaluation. We find that OMP+MC achieves the best result, with a AUC of 0.5433. This is because that the high response in the code means there is a strong connection between the testing feature with some atoms in the dictionary. This happens when the features have the similar pattern as those atoms convey. Therefore, the high response also implies that the testing feature is normal. However, we also notice that NC detection, which also considers the non-zeros distribution in sparse codes, performs relative poorly. This may be due to the type of dictionary being adopted, or due to the principle of how the OMP code is generated, which are based on the reconstruction error of the chosen atoms, rather than the concentrated atoms.

\begin{figure}[!htb]
\vskip 0.2in
\begin{center}
\centerline{\includegraphics[width=0.95\columnwidth]{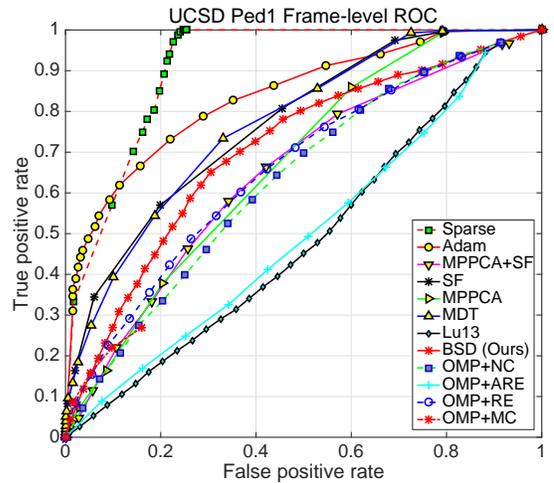}}
\caption{Comparison with state-of-the-art abnormality detection approaches.}
\label{fig:detection_stoa}
\end{center}
\vskip -0.2in
\end{figure}

Finally we compare combining OMP codes and various detection methods with state-of-the-art abnormality detection algorithms. Comparison of AUC in the frame-level evaluation of UCSD Ped1 is shown in Fig.~\ref{fig:detection_stoa}, and quantized evaluations are shown in Tab.~\ref{stoa} Compared with state-of-the-art algorithms, combining OMP codes with detection methods outperforms other methods on two criteria evaluation, which verifies the effectiveness of sparse codes generated by greedy algorithms; furthermore, maximum coordinate detection outperforms other methods, which implies that a high response (large code value) could contribute to the detection.

\begin{table*}
\caption{Comparative values of AUC, EER and EDR on UCSD Ped1 dataset.}
\label{stoa}
\begin{center}
\begin{tabular}{|l|c|c|c|c|c||}
\hline
Method & AUC (frame-level) & EER & AUC (pixel-level)  &  EDR\\
\hline\hline
SF-MPPCA~\cite{Mahadevan2010} & 0.5900 & 0.3200 & 0.2130 & 0.3200\\
MDT~\cite{Mahadevan2010} & 0.8180 & 0.2500 & 0.4610 & 0.2500\\
Lu13\cite{Lu13} & 0.5842 & 0.4413 & 0.3622 & 0.5826 \\
OMP+RE  &  \textbf{0.6603} &  \textbf{0.3823} & 0.5386 & 0.5113 \\
OMP+ARE  & 0.5013 & 0.5081 & 0.5317 & 0.5113 \\
OMP+NC & 0.5697 & 0.5055 & 0.5397 & 0.5113 \\
OMP+MC  & 0.6339 & 0.4016 &  \textbf{0.5433} &  \textbf{0.5113} \\
\hline
\end{tabular}
\end{center}
\end{table*}

\section{Discussions and Conclusion}\label{sec:conclusion}
In this paper, we give a comprehensive study of sparse codes, in respect to their performance in abnormal event detection. We compare two category sparse codes: codes generated by greedy algorithms and those generated by L1-norm solutions. Various aspects are covered: computational cost, recovery ability, sparsity, and their detection performance. Furthermore, we explore into the sparse codes, and compare different methods to determine whether a testing code is an anomaly or not.

Experimental results show that greedy algorithms can obtain good detection results with fewer computations. Among the top three best detection results, two are greedy algorithms. Considering the computation requirement, which limits some L1-norm algorithms from being applied in real surveillance applications, greedy algorithms are promising. When combining OMP codes with various detection measurements, maximum coordinate measurement outperforms other methods, which implies that the high response in the code could help the detection result.

We have also found that due to the large amount of video data only the OMP code is acceptable, which is mainly for computational reasons. Despite the great progress being made in the optimization field, the applicability of various optimization solutions is still unknown. Therefore, one line of future work could focus on more practical sparse code algorithms. Another line of work may fall into discriminative feature selection to reduce the computation of sparse codes generation.

\bibliographystyle{IEEEbib}
\bibliography{icme2016template}

\begin{thebibliography}{10}

\bibitem{donoho2002optimally}
D.~L. Donoho and M.~Elad,
\newblock {\em Optimally Sparse Representation in General (non-orthogonal)
  Dictionaries Via L1 Minimization},
\newblock Department of Statistics, Stanford University, 2002.

\bibitem{JiangTPAMI}
Z.~Jiang, Z.~Lin, and L.~S. Davis,
\newblock ``Label consistent k-svd: Learning a discriminative dictionary for
  recognition,''
\newblock {\em PAMI}, vol. 35, no. 11, pp. 2651--2664, 2013.

\bibitem{Huang06sparserepresentation}
K.~Huang and Selin Aviyente,
\newblock ``Sparse representation for signal classification,''
\newblock in {\em In Adv. NIPS}, 2006.

\bibitem{MP1993}
S.~G. Mallat and Z.~Zhang,
\newblock ``Matching pursuits with time-frequency dictionaries,''
\newblock {\em Signal Processing, IEEE Transactions on}, vol. 41, no. 12, pp.
  3397--3415, Dec 1993.

\bibitem{OMP1993}
Y.~C. Pati, R.~Rezaiifar, and P.S. Krishnaprasad,
\newblock ``Orthogonal matching pursuit: recursive function approximation with
  applications to wavelet decomposition,''
\newblock in {\em Signals, Systems and Computers}, Nov 1993, pp. 40--44 vol.1.

\bibitem{Chen98atomicdecomposition}
S.~Chen, D.~L. Donoho, and M.~A. Saunders,
\newblock ``Atomic decomposition by basis pursuit,''
\newblock {\em SIAM Journal on Scientific Computing}, vol. 20, pp. 33--61,
  1998.

\bibitem{Aharon05k}
A.~Michal, E.~Michael, and B.~Alfred,
\newblock ``K-svd: Design of dictionaries for sparse representation,''
\newblock in {\em SPARS'05}, 2005, pp. 9--12.

\bibitem{RenBMVC2015}
H.~Ren, W.~Liu, S.~Escalera S.~Olsen, and T.~B. Moeslund,
\newblock ``Unsupervised behavior-specific dictionary learning for abnormal
  event detection,''
\newblock in {\em BMVC}, 2015.

\bibitem{Lu13}
C.~Lu, J.~Shi, and J.~Jia,
\newblock ``Abnormal event detection at 150 fps in matlab,''
\newblock in {\em ICCV}, 2013, pp. 2720--2727.

\bibitem{zhang2009}
T.~Zhang,
\newblock ``Some sharp performance bounds for least squares regression with l1
  regularization,''
\newblock {\em Ann. Statist.}, vol. 37, pp. 2109--2144, 2009.

\bibitem{StOMP2012}
D.~L. Donoho, Y.~Tsaig, I.~Drori, and J.~L Starck,
\newblock ``Sparse solution of underdetermined systems of linear equations by
  stagewise orthogonal matching pursuit,''
\newblock {\em Information Theory, IEEE Transactions on}, vol. 58, no. 2, pp.
  1094--1121, 2012.

\bibitem{Focuss2001}
J.F. Murray and K.~Kreutz-Delgado,
\newblock ``An improved focuss-based learning algorithm for solving sparse
  linear inverse problems,''
\newblock in {\em Signals, Systems and Computers}, Nov 2001, vol.~1, pp.
  347--351 vol.1.

\bibitem{tibshirani96regression}
R.~Tibshirani,
\newblock ``Regression shrinkage and selection via the lasso,''
\newblock {\em Journal of the Royal Statistical Society (Series B)}, vol. 58,
  pp. 267--288, 1996.

\bibitem{Mahadevan2010}
V.~Mahadevan, W.~Li, V.~Bhalodia, and N.~Vasconcelos,
\newblock ``Anomaly detection in crowded scenes,''
\newblock in {\em CVPR}, 2010, pp. 1975--1981.

\bibitem{KratzN09}
L.~Kratz and K.~Nishino,
\newblock ``Anomaly detection in extremely crowded scenes using spatio-temporal
  motion pattern models,''
\newblock in {\em CVPR}, 2009, pp. 1446--1453.

\end{thebibliography}

\end{document}